\documentclass[10pt,twocolumn,letterpaper]{article}

\usepackage{ICCV2025-Author-Kit-Feb/iccv}              
\definecolor{iccvblue}{rgb}{0.21,0.49,0.74}
\usepackage[pagebackref,breaklinks,colorlinks,allcolors=iccvblue]{hyperref}

\def\paperID{9} 
\def\confName{ICCV}
\def\confYear{2025}

\title{Fourier Domain Adaptation for Traffic Light Detection in Adverse Weather*}

\author{Ishaan Gakhar\\
Manipal Institute of Technology,\\
Manipal Academy of Higher Education, \\
Manipal, India\\
{\tt\small ishaangakhar04@gmail.com}
\and
Aryesh Guha$\dagger$\\
Dept. of Electrical and Electronics Engineering,\\
Manipal Institute of Technology,\\
Manipal Academy of Higher Education, \\
Manipal, India\\
{\tt\small aryeshguha1510@gmail.com}
\and
Aryaman Gupta$\dagger$\\
Manipal Institute of Technology,\\
Manipal Academy of Higher Education,\\
Manipal, India\\
{\tt\small aryamangupta004@gmail.com}
\and
Amit Agarwal\\
Enterprise Analytics $\&$ Data Science Artificial Intelligence,\\
Center of Excellence Wells Fargo International Solutions Private Limited.\\
{\tt\small aagarwal3@cs.iitr.ac.in}
\and
Ujjwal Verma\\
Department of Electronics and Communication Engineering,\\
Manipal Institute of Technology,\\
Manipal Academy of Higher Education,\\
Manipal, India\\
{\tt\small ujjwal.verma@manipal.edu}
}

\begin{document}
\maketitle
\begin{abstract}

Traffic light detection under adverse weather conditions remains largely unexplored in ADAS systems, with existing approaches relying on complex deep learning methods that introduce significant computational overheads during training and deployment. This paper proposes Fourier Domain Adaptation (FDA), which requires only training data modifications without architectural changes, enabling effective adaptation to rainy and foggy conditions. FDA minimizes the domain gap between source and target domains, creating a dataset for reliable performance under adverse weather.

The source domain merged LISA and S²TLD datasets, processed to address class imbalance. Established methods simulated rainy and foggy scenarios to form the target domain. Semi-Supervised Learning (SSL) techniques were explored to leverage data more effectively, addressing the shortage of comprehensive datasets and poor performance of state-of-the-art models under hostile weather.

Experimental results show FDA-augmented models outperform baseline models across mAP50, mAP50-95, Precision, and Recall metrics. YOLOv8 achieved a 12.25\% average increase across all metrics. Average improvements of 7.69\% in Precision, 19.91\% in Recall, 15.85\% in mAP50, and 23.81\% in mAP50-95 were observed across all models, demonstrating FDA's effectiveness in mitigating adverse weather impact. These improvements enable real-world applications requiring reliable performance in challenging environmental conditions.
\end{abstract}    
\section{Introduction}
\label{sec:intro}

\footnote{$\dagger$ indicates equal contribution. *The views expressed in this article are of the authors and do not represent the views of Wells Fargo. This article is for informational purposes only. Nothing contained in the article should be constructed as investment advice. Wells Fargo makes no express or implied warranties and expressly disclaims all legal, tax and accounting implications related to this article.}

Environmental perception, along with behaviour decision and motion control, is an essential task in the domain of autonomous driving. With artificial intelligence and computer vision techniques playing integral roles in autonomous driving, object detection plays a crucial role in understanding surrounding environmental scenarios \cite{xu2022opv2vopenbenchmarkdataset,zhao2020fusion}. Adverse weather conditions compromise sensor accuracy and real-time performance, resulting in image degradation issues. These weather anomalies affect the object detection performance of autonomous vehicles \cite{qiu2023idod,wang2019pso}.

Detecting and recognizing traffic lights is important for ensuring safety in the field of autonomous driving. Vision-based traffic light detection and recognition in traffic scenes is still a challenge \cite{zang2019impact,Bijelic_2020_CVPR} to be successfully overcome by the autonomous driving industry, especially when there are disturbances due to weather conditions that contribute to the degradation in the performance of object detection models. State-of-the-art (SOTA)  object detection algorithms, like YOLOv5 \cite{yolov5}, YOLOv6 \cite{li2023yolov6}, YOLOv8 \cite{yolov8_ultralytics} and YOLOv10 \cite{wang2024yolov10realtimeendtoendobject}, perform well in clear weather conditions. However, when used in detrimental weather conditions, such as environments with rain or fog, a significant deterioration in their performance is observed, as mentioned in \cite{s23208471}. Consequently, there is a pressing need for methods that enable these models to adapt to the variability introduced by challenging weather conditions, ideally without relying on annotated labels as a prerequisite.

Unsupervised Domain Adaptation (UDA) is a form of transfer learning that involves adapting a model trained on one domain that has an abundance of annotations (source domain) to demonstrate results in a related yet distinct domain (target domain) characterized by a different data distribution that lacks annotations. Training the model solely within the source domain fails to produce satisfactory outcomes owing to the occurrence of covariate shift, as mentioned in \cite{yang2020fda}. The primary objective of domain adaptation techniques is to facilitate knowledge transfer from a well-labelled source domain to a target domain that lacks labels. This process is essential as minor alterations in low-level statistics notably impact the model's effectiveness. This is particularly relevant for leveraging existing datasets with annotations to enhance performance on related tasks that lack such annotations, like Advanced Driver Assistance Systems (ADAS), security and surveillance systems, etc. Domain adaptation thus plays a pivotal role in addressing the challenge of data scarcity in supervised learning scenarios by bridging the discrepancy between different but related domains. 

This work proposes Fourier Domain Adaptation (FDA) \cite{yang2020fda}, a non-parametric domain adaptation algorithm. Here, the Fast Fourier Transform (FFT) of each input image (clear weather) in the source domain is computed along with the FFT of a randomly chosen target image (rainy or foggy). Then, the low-level frequencies of the target image replaces the low-level frequencies of the source image, before the images are reconstructed using Inverse FFT (iFFT). The source domain is formed by merging two widely used datasets, LISA \cite{jensen2016vision} and SJTU: Small Traffic Light Dataset (S\textsuperscript{2}TLD) \cite{yang2022scrdet++}, and employing various augmentations to address the class imbalance against images containing yellow traffic lights. By Leveraging pre-existing techniques for fog addition \cite{tran2024toward} and rain addition \href{https://github.com/ShenZheng2000/Rain-Generation-Python}{(https://github.com/ShenZheng2000/Rain-Generation-Python)}, the target domain for traffic light detection in rainy and foggy scenarios was created. 
This enables models to learn domain-invariant features, enhancing their ability to detect traffic lights compared to models trained on clear weather conditions as observed.
State-of-the-art UDA algorithms often entail training a deep neural network, which undergoes difficult adversarial training to make the model invariant in the binary selection of source/target domain \cite{yang2020fda} whereas FDA requires no trainable parameters or complex architectures. It facilitates the learning of domain-invariant features by introducing variability, which helps the model learn high-level characteristics without relying on low-level characteristics from the sensor, the illuminant, or other sources of low-level variability. Therefore, this non-parametric, traditional signal processing algorithm can be used as an alternative to dense deep-learning based, computationally heavy, parametric state-of-the-art unsupervised domain adaptation (UDA) algorithms \cite{pmlr-v80-hoffman18a}. This serves as motivation to apply FDA to weather-affected images. The source dataset was benchmarked using various state-of-the-art methods, and their performance was qualitatively and quantitatively evaluated across different metrics. 

In order to tackle the challenge posed by the limited availability of annotated datasets for traffic light detection in hostile weather, semi-supervised learning (SSL) techniques were employed. SSL, which leverages both labelled and unlabeled data, is particularly advantageous in situations where obtaining a comprehensive labelled dataset is costly or logistically difficult. By integrating a smaller set of accurately labelled traffic light images with a larger pool of unlabeled images, SSL allows leveraging the already limited data and demonstrates that even 50\% of the available data can be used to produce comparable results. This approach aims to mitigate data scarcity inherent in the domain. Through the application of semi-supervised learning, it becomes feasible to develop more effective algorithms that can accurately identify and classify traffic lights in diverse driving conditions at much less cost, ultimately contributing to advancements in intelligent transportation systems.
Hence, the contributions in this work can be summarized as:
 \begin{itemize}
     \item Proposing and demonstrating the effectiveness of a non-parametric method of Fourier Domain Adaptation for Traffic Light detection in adverse weather conditions requiring no pretraining or dense architectures.
     \item Leveraging Semi-Supervised Learning in the realm of Traffic Light Detection as a potential solution to the lack of datasets in the domain and demonstrating comparable performance with only 50\% labelled data.
     \item Combining widely used datasets and mitigating class imbalance to synthetically generate realistic rainy and foggy images employing effective pre-existing methods.
 \end{itemize}

\section{Related Works}
\label{sec:RelatedWorks}

\begin{figure*}[t]
    \small
    \centering
    \includegraphics[width=\textwidth, height=3in, keepaspectratio]{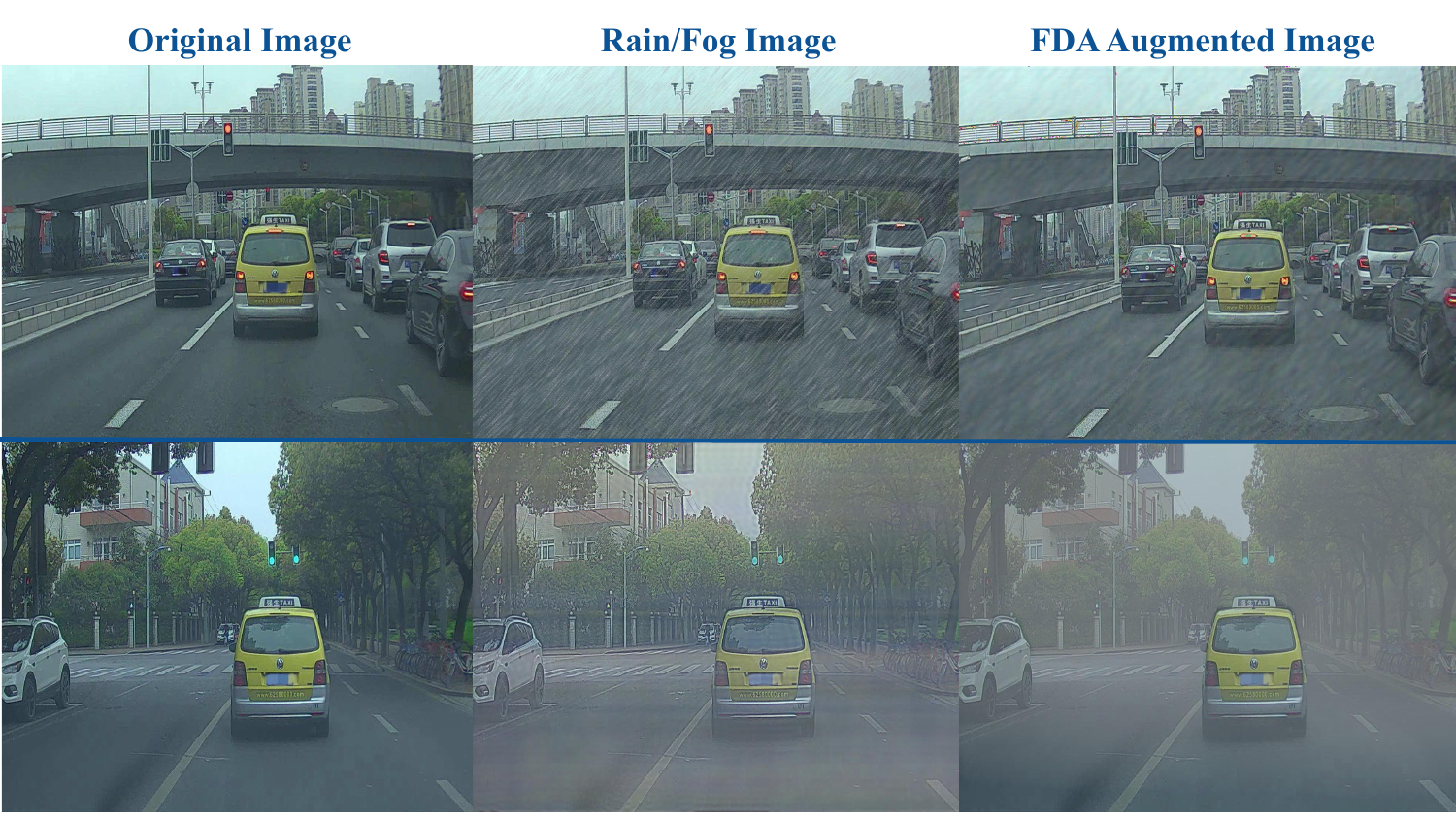}
    \caption{Impact of FDA on Rainy and Noisy images. The first column contains original images, the 2nd column adds rain/fog to the corresponding image and the 3rd column presents the effects of FDA on the images of the 2nd column. Each row represents a different instance from the dataset. As evident, applying FDA to the images increases visibility and the ability to detect the traffic light.}
    \label{fig:FDA_Output}
 \end{figure*}

\subsection{Object Detection}
In recent years, significant advancements have been made in object detection, driven by the development of deep learning-based techniques. One of the most influential works is the YOLO (You Only Look Once) series, introduced by \cite{redmon2016lookonceunifiedrealtime}, which revolutionized object detection by achieving real-time performance while maintaining high accuracy. YOLOv3 \cite{redmon2018yolov3} and its successors, YOLOv5 \cite{yolov5} and YOLOv8 \cite{yolov8_ultralytics}, have further improved detection capabilities by incorporating advanced features and optimization techniques. Additionally, Mask R-CNN by He et al. \cite{he2017mask} extended Faster R-CNN by adding a branch for predicting segmentation masks, thereby providing a unified approach for object detection and instance segmentation. A novel object detection model that leverages transformers instead of traditional CNNs was introduced in \cite{carion2020endtoendobjectdetectiontransformers}. It simplifies the object detection pipeline by directly predicting bounding boxes and class labels from image pixels, eliminating the need for anchor boxes and non-maximum suppression. 

\subsection{Object Detection in Unfavourable Weather}

Object detection in hostile weather conditions has been a topic of recent interest to improve the effectiveness of autonomous vehicles and surveillance systems. The use of YOLOv5 for real-time identification of cars, traffic lights, and pedestrians in various weather conditions, including challenging scenarios like rain and fog, is explored in \cite{electronics11040563}. A novel approach has been used in \cite{a13110271} combining visible and infrared image features using a CNN. By fusing these complementary features, the method enhances the accuracy of detection in low-visibility conditions.
A network based on the joint optimal learning of an image-defogging module (IDOD) and YOLOv7 detection modules is proposed in \cite{qiu2023idod}. This approach is designed for low-light foggy images, where the SAIP module's parameters are predicted by a separate CNN network, and the AOD module performs defogging by optimizing the atmospheric scattering model.
The model proposed in \cite{JI2024100167} significantly improves the metrics but focuses solely on detecting traffic signs rather than lights. Given that lights have higher intensity, their varying brightness levels can introduce difficulty in distinguishing shapes and colors, which may affect detection accuracy.

\subsection{Traffic Light Detection}

The detection and recognition of traffic lights are crucial components for the safety and efficiency of autonomous vehicles and intelligent transportation systems. In \cite{8569575}, the authors first passed the images through a deep neural object detection architecture to get traffic light candidates and then used a reward-based system where each detected traffic light earns rewards concerning features extracted from both spatial and temporal domains to reduce false positives. One drawback noted was that the classifier performed poorly for yellow lights, likely due to fewer training examples for them. To detect traffic lights at night and at a distance, the paper by Phuc et al. \cite{9268343} used an approach combining the benefits of hand-crafted features and deep learning techniques. However, this method fails to detect traffic lights in complex regions with many traffic light-like objects. An improved YOLOv4 with a lightweight ShuffleNetv2 backbone, along with an enhanced K-Means clustering algorithm for bounding boxes and a novel CS2A mechanism, was introduced by \cite{s22207787} for traffic light recognition on mobile devices. These updates, combined with data augmentation techniques, improve recognition accuracy, reduce model size, and enhance detection speed and small target recognition. At the same time, though, the overall mAP is limited due to a small dataset, and the model cannot generalize very well due to the poor fitting ability of YOLOv4.

\subsection{Fourier Domain Adaptation (FDA) for Classification}

Domain adaptation is a crucial technique in computer vision, designed to bridge the gap between source and target domains with different distributions. Traditional methods, such as those developed by Tzeng et al .\cite{Tzeng_2017_CVPR} and Long et al. \cite{8454781}, have significantly advanced the field by focusing on feature alignment and adversarial training.
An extension of domain adaptation is FDA, which aligns the spectral properties of images through the Fourier transform. Yang and Soatto \cite{yang2020fda} pioneered this approach for cross-domain semantic segmentation. This method has been further explored in various applications; for example, in \cite{app14093727}, the authors have used the concept of swapping the low-frequency part of the source and target domains used in FDA using the concept of Farthest Point Sampling (FPS).

\subsection{Semi-Supervised Learning for Classification}

Previous work has addressed object detection under adverse weather conditions \cite{s23208471} \cite{10715515}. However, to the best of our knowledge, no datasets specifically for traffic light detection under adverse weather conditions exist at the time of writing. In this context, semi-supervised learning becomes increasingly important as it leverages both labelled and unlabelled data, addressing the challenge of manually annotating large datasets. To address this issue for the more general domain of traffic signs, the authors have used a combination of weakly supervised and semi-supervised learning to address the problems of imbalanced data and optimizing pseudo-labels generated on unlabelled data. A ranking-based pseudo-label generation system has been employed in \cite{xu2022opv2vopenbenchmarkdataset}, where certain objects are given more importance than others in the object detection system leading to an improvement in performance over supervised learning methods. 


\section{Methodology}

\begin{figure*}[t]
    \centering
    \small
    \includegraphics[width=\textwidth, height=3.7in]{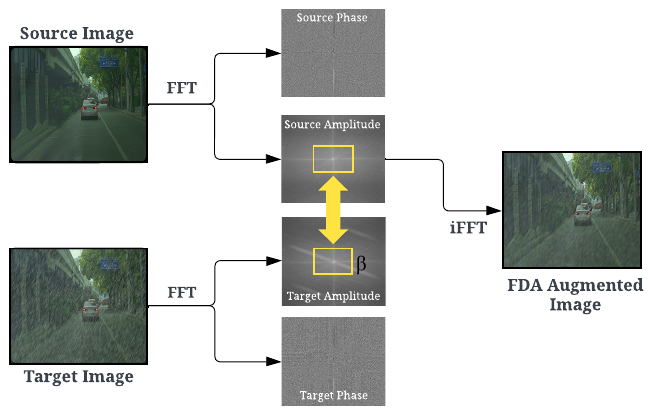}
    \caption{Shown above is the utilization of a target image to provide the desired style to a source image. The technique achieves this by swapping the low-frequency component of the source image (upper) with that of the target image (lower). The resulting FDA-augmented image exhibits a reduced perceptual domain gap, which helps object detection models adapt better to adverse weather conditions, as mentioned in the benchmarking section. It is important to note that the phase  (high-frequency component) and amplitude (low-frequency component) images retain the same resolution as the original images; they are depicted smaller in the diagram solely for visual representation.}
    \label{fig:fda4}
\end{figure*}


\label{sec:Methodology}

This work aims to overcome the effects of adverse weather conditions and the lack of datasets for such applications. To this end, two widely used datasets were combined and modified, followed by Fourier Domain Adaptation to modify the spectral signals obtained by using the Fast-Fourier Transform (FFT). The spectral signals are then converted back to the image space using Inverse Fast-Fourier Transform (iFFT). This produces images that consist of high-level characteristics of the source domain (clear weather) while incorporating the low-level characteristics of the target domain (rainy or foggy conditions). 

Fourier Domain Adaptation (FDA) \cite{yang2020fda} is a non-parametric technique that leverages Fast Fourier transforms to improve domain adaptation by modifying the frequency components of images. The key idea is to replace the low-frequency components of a source domain image (e.g., clear weather) with those from a target domain image (e.g., rainy or foggy conditions). Since low-frequency components encode global structural information while high-frequency components capture finer details, this transformation helps the model retain structural consistency while adapting to new environmental conditions. By training models with FDA-augmented data and labels of the source domain (target domain labels are absent), the model becomes more robust to domain shifts caused by adverse weather conditions. This method is especially helpful for mitigating the effects of rain or fog, as it can be challenging to find data with these conditions along with proper annotations. Rain and fog primarily cause blurring, which reduces visibility. FDA curbs this issue by exposing the models to frequency-adjusted samples during training. This approach helps the models generalize more effectively to real-world conditions.

\begin{equation}
D^{s} = \{(x^s_i, y^s_i) \sim P(x^s, y^s)\}_{i=1}^{N_s}, \quad  x^s \in {R}^{H \times W \times 3}
\label{eq:one}
\end{equation}

As defined in (\ref{eq:one}), the source dataset with clear weather conditions, \(D^{s}\), is utilized, where \(x^{s}\) represents colour images, and \(y^{s}\) denotes the corresponding bounding box labels. The dataset \(D^{s}\) consists of \(N_s\) such image-label pairs, each sampled from the joint probability distribution \(P(x^{s}, y^{s})\). Similarly, the target dataset is defined as \(D^t = \{ x^t_i \}_{i=1}^{N_t}\), which contains target images with adverse weather conditions and no ground truth bounding box labels. Here, \(H\) and \(W\) represent the height and width of the image, respectively.

This transformation is implemented by (Fast) Fourier Transform (FFT) as in \cite{frigo1998fftw}. \begin{math}\mathcal{F}\end{math} represents the Fourier transform, which converts an image into spectral components (phase and amplitude), whereas \begin{math}\mathcal{F}\end{math}\textsuperscript{-1} represents the inverse Fourier transform, converting spectral signals back into the image space.

The Fourier Transform \begin{math}\mathcal{F}\end{math} for an image x with a single channel, gives:
\begin{equation}
    \mathcal{F}(x)(m, n) = \sum_{h,w} x(h, w) e^{-j2\pi \left( \frac{h}{H} m + \frac{w}{W} n \right)}, \quad j^2 = -1
\end{equation}  
Here, $h$ and $w$ denote pixel indices in the vertical and horizontal directions, respectively, while $m$ and $n$ represent the corresponding frequency coordinates in the Fourier domain. 
Since low-frequency components correspond to the central region of the frequency spectrum, we define a binary mask $M_\beta$ that selects this central low-frequency area as:

\begin{equation}
M_\beta(h, w) = \mathbf{1}_{(h, w) \in [-\beta H : \beta H, -\beta W : \beta W]}
\end{equation}

The origin (0,0) is taken to be at the center of the image in the frequency domain. The parameter $\beta \in (0, 1)$ controls the size of this low-frequency region as a proportion of the image dimensions and is thus resolution-independent. This makes Fourier Domain Adaptation (FDA) applicable even when the source and target images differ in resolution.
Given two randomly sampled images from the source and target pair, \( x\textsuperscript{s} \sim D\textsuperscript{s} \) (clear weather) and \( x\textsuperscript{t} \sim D\textsuperscript{t} \) (adverse weather), Fourier Domain Adaptation can be formalized as:
\begin{equation}
\begin{split}
x^{s \rightarrow t} &= \mathcal{F}^{-1}\Big[ M_\beta \circ \mathcal{F}^A(x^t) \\
&\quad + (1 - M_\beta) \circ \mathcal{F}^A(x^s), \mathcal{F}^P(x^s) \Big]
\end{split}
\end{equation}


Where $\mathcal{F}^A, \mathcal{F}^P:$  represent the amplitude and phase components of the Fourier transform  $\mathcal{F}$ of the corresponding image.

The low-frequency amplitude components of the source image \( x^s \) are replaced with those from the target image \( x^t \), showing adverse weather. The modified spectral representation of \( x\textsuperscript{s} \), with its phase component retained, is then transformed back into the image \( x^{s \to t} \) using iFFT. This resultant image \( x^{s \to t} \) retains the content of \( x\textsuperscript{s} \) but adopts the appearance of a sample from \( D\textsuperscript{t} \). 

This process is illustrated in Fig. \ref{fig:fda4}, where the mask \( M_{\beta} \) has been depicted. The optimal $\beta$ value chosen for the FDA was 0.15. This value was used to construct the adapted dataset $D^{s \to t}$. The final merged dataset containing the FDA-augmented images and the ground truth bounding box labels of source dataset $D\textsuperscript{s}$ is used to train various state-of-the-art object detection models, such as YOLOv5, YOLOv6, and YOLOv8. The effect of FDA is illustrated in Fig. \ref{fig:FDA_Output} and Fig. \ref{fig:FDA_comparisonl}.

As discussed in Section \ref{sec:intro}, there exists a lack of datasets for traffic light detection in hostile weather. To validate the proposed method for learning domain-invariant features, we merged two widely used datasets: LISA and S\textsuperscript{2}TLD. Here, a severe class imbalance against yellow lights was noted, and augmentations were applied to increase the proportion of yellow lights in the dataset. Experiments were conducted with varying proportions and several models to choose an optimal percentage, to which rain and fog were added artificially. Furthermore, semi-supervised learning was leveraged to demonstrate comparable results with only 50\% labelled data. These details are provided in the following section.

The details regarding the choice of hyperparameters for the addition of fog and rain, their effects and the details regarding convergence are mentioned in the supplementary material.

\section{Experiments}
    
\label{sec:Experiments}

\begin{figure*}[t]
    \centering
    \small
    \includegraphics[width=\textwidth]{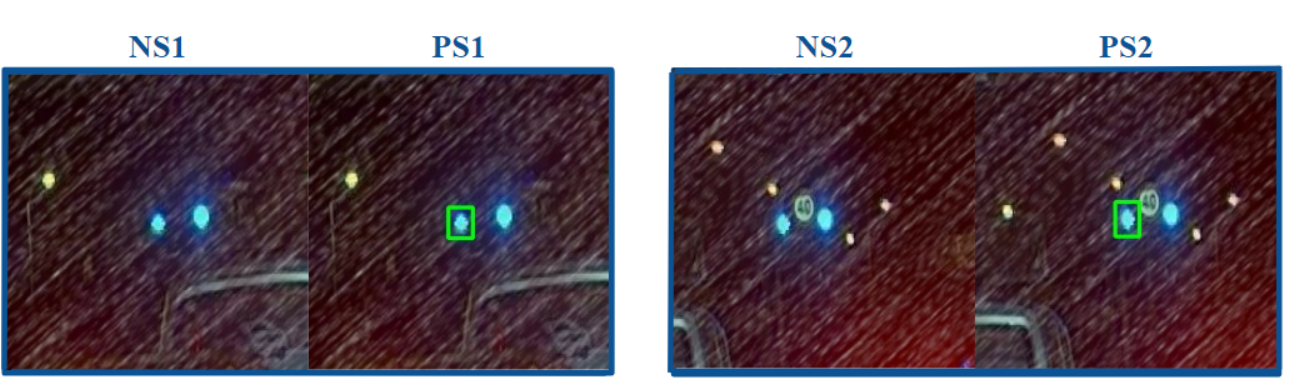}
    \caption{Qualitative comparison on the $D\textsuperscript{t}$ dataset for detections from YOLOv8 under adverse weather conditions. The images NS1 and NS2 represent negative samples where the model trained only on the $D\textsuperscript{s}$ dataset failed to detect the traffic lights. In contrast, PS1 and PS2 are the same positive samples where the model trained on the FDA-augmented dataset \( D^{s \to t} \)  successfully detected the traffic lights. This highlights how FDA improves the model’s robustness by making it better suited to handle challenging visual conditions such as rain or fog.}
    \label{fig:light_instances}
\end{figure*}

To validate our method, two widely used datasets - S\textsuperscript{2}TLD and LISA were merged for experiments.

\textbf{The Shanghai Jiao Tong University (SJTU): Small Traffic Light Dataset (S\textsuperscript{2}TLD) }\cite{yang2022scrdet++} is a dataset designed for traffic light detection and automotive navigation tasks. It contains images with resolutions of approximately 1920 x 1080 pixels and 1280 x 720 pixels. The dataset consists of 14,130 instances categorized into five classes: red, yellow, green, off, and wait on of which only the red, green, and yellow classes were used. 

\begin{table}[h]
    \centering
    \small
    \caption{Brief summarization of the individual datasets chosen. The S$^2$TLD is a multiclass dataset, which is why its sum exceeds 100\%. As evident, there innately exists a severe class imbalance against the 'Yellow' class.}
    \label{table:datasets_percentages}
    \begin{tabular}{|c|c|c|c|c|}
    \hline
        \textbf{Dataset} & \textbf{\#Images} & \textbf{\#Red} & \textbf{\#Green} & \textbf{\#Yellow} \\ \hline \hline LISA & 38323 & 48.51\% & 48.60\% & 2.88\%\\ \hline
        S\textsuperscript{2}TLD & 5153 & 73.29\% & 56.88\% & 3.41\%\\ \hline \hline
    \end{tabular}
\end{table}

\textbf{LISA Traffic Light Dataset} 
\cite{jensen2016vision} is a comprehensive resource for traffic light recognition (TLR) tasks used for Autonomous Driving Systems. It consists of images of 14 classes of the resolution 1280 x 960. It contains annotated traffic lights (of which only the ones containing red, yellow, and green were used), captured using a stereo camera mounted on a vehicle. 

FDA employs a hyperparameter ($\beta$), which determines the size of the spectral neighbourhood swapped between the source and target images. $\beta$ = \{0, 0.05, 0.10, 0.15\} were utilized for experiments. Higher values of $\beta$ tend to create artefacts, as mentioned in \cite{yang2020fda}. As explained in Section \ref{sec:Methodology}, $\beta$ is independent of the resolution of images, which allows this method to be adapted to images of various sizes and crops. FDA was performed on the source domain images using the target domain images to generate a modified dataset, \(D^{s \to t}\).

For experiments, the final dataset was split into 70\% for training, 20\% for validation, and 10\% for testing. All the images are resized to the resolution of 1280 x 1080 using bicubic interpolation, and the labels are adjusted accordingly. The training set consisted of images from  \(D^{s \to t}\), the FDA-augmented dataset, while the validation and test sets used images from \(D^t\), the target domain with rain and fog effects.

Furthermore, we explore Semi-Supervised Learning (SSL) to mitigate the lack of datasets available by initially training YOLOv6m on 50\% of the annotated training data to generate pseudo-labels for the remaining unlabeled half. This new compiled dataset of 50\% original labels, and 50\% pseudo-labels is used to benchmark models. A confidence threshold of 50\% was applied to filter the predicted labels, ensuring that only high-confidence detections were retained. 

To address the class imbalance evident in Table \ref{table:datasets_percentages}, various proportions of yellow traffic light images were evaluated to identify the optimal percentage for benchmarking. The models were trained on datasets with various percentages of yellow images, including 5\%, 7\%, 9\%, 11\%, and 13\%. These experiments indicate that the dataset comprising 13\% yellow images led to convergence, suggesting that generalizability across all models is achieved largely at this percentage, making it suitable for the source domain. Detailed results of these experiments, along with the augmentations performed in order to mitigate class imbalance, are provided in the supplementary material.

\section{Results}
\label{results}

\begin{figure}[t]
    \centering
    \small   
    \includegraphics[width=0.47\textwidth, height=3.25in]{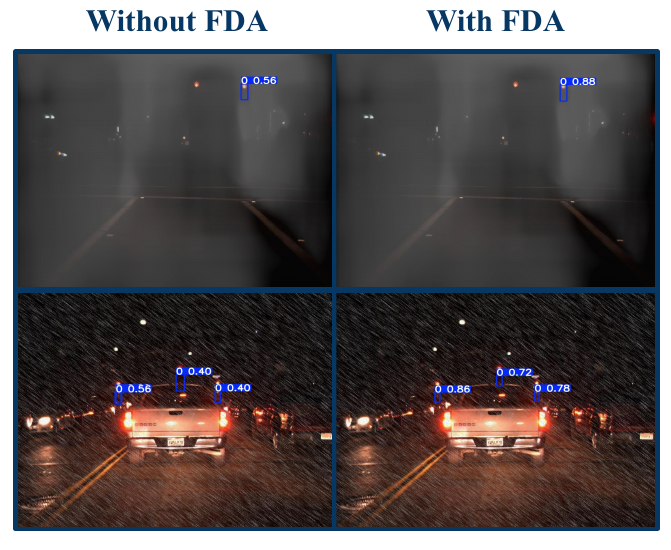}
    \caption{Comparisons of outputs from inferring on YOLOv8m. The first image in each row is the output of the model trained on D$^s$, and the second is the output when trained on D$^{s\rightarrow t}$. Each bounding box has the label (0 implies Red) and the associated confidence score.}
    \label{fig:FDA_comparisonl}
\end{figure}

The performance of the proposed approach is evaluated against confidence-dependent and confidence-independent metrics like Precision, Recall, mAP50, and mAP50-95. Here, the qualitative and quantitative results are presented.

\begin{figure}[t]
    \centering
    \small
    \includegraphics[width=0.47\textwidth, keepaspectratio]{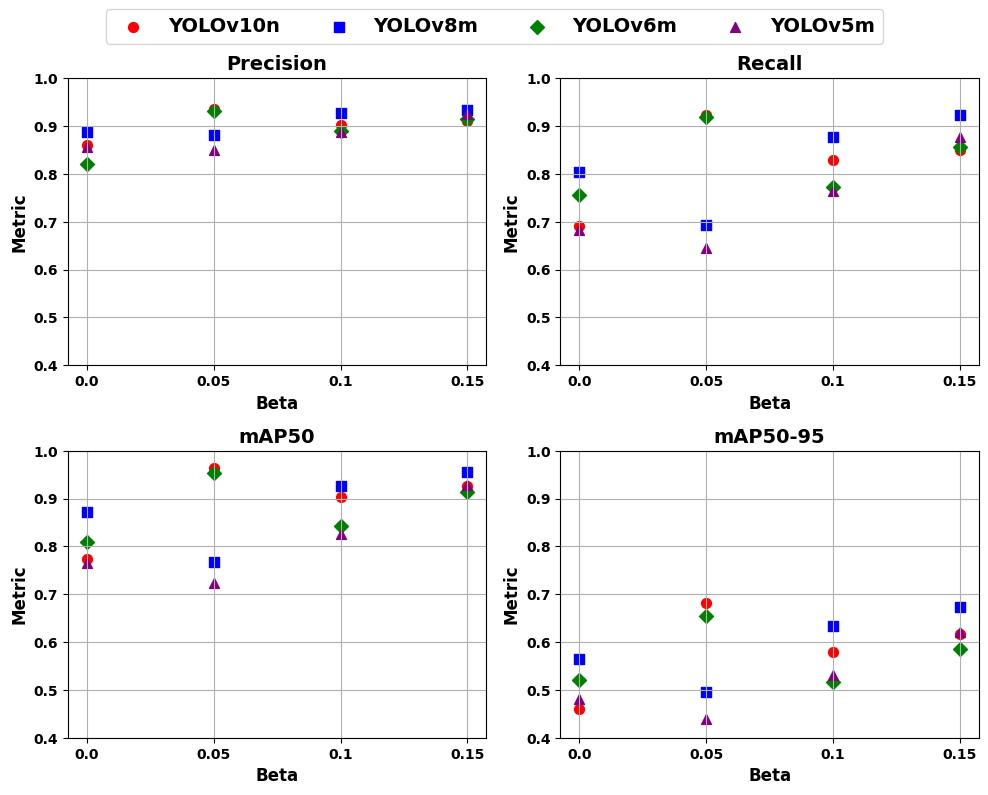}
    \caption{Comparisons of results of various models with different $\beta$ values. It is observed that these models perform best and yield the most consistent results at $\beta$ = 0.15.}
    \label{fig:results_graph}
\end{figure}

As demonstrated in Table \ref{table:classwise_results}, the class imbalance has been mitigated by incorporating augmented images. The models exhibit excellent performance, especially in detecting yellow lights across various degrees of FDA, as indicated by the parameter $\beta$, demonstrating the effectiveness of the proposed methodology. For $\beta = 0.15$, the average percentage increase in metrics from training on D$^s$ to D$^{s\rightarrow t}$ for mAP50 and mAP50-95 for the yellow class is 16.12\% in YOLOv8, 22.63\% in YOLOv6, and 31.80\% in YOLOv5, respectively.

As shown in Fig. \ref{fig:results_graph}, when $\beta$ = 0 (unchanged source domain), models trained on \(D^s\) and inferred on \(D^t\) exhibit sub-par performances. This degradation in metrics highlights how exhaustive experimentation ensures the addition of not only effective but also realistic rain and fog. It simulates real-world weather conditions in the label-less target domain dataset, thereby validating the observed performance drop and pointing to its real-world applications. Performance across all metrics improves with the increase in $\beta$, where the best results are observed at $\beta$ = 0.15. According to Fig. \ref{fig:results_graph}, results at $\beta$ = 0.15 are higher and most tightly clustered, showing better performance and consistency across metrics, with YOLOv8 displaying the highest performance metrics - Precision of 0.93, Recall of 0.92, mAP50 of 0.96, and mAP50-95 of 0.67. An average increase of \textbf{12.25\%} across all metrics for YOLOv8 was observed when training on D$^{s\rightarrow t}$. The authors believe YOLOv8m outperforms YOLOv10n because the nano version is parametrically smaller, leading to reduced performance on complex features. However, the nano version was used as it is less computationally intensive compared to the bulkier medium version. The average increase across all metrics, across all models is observed to be \textbf{16.81\%}. On average, percentage increases of 7.69\% in Precision, 19.91\% in Recall, 15.85\% in mAP50, and 23.81\% in mAP50-95 is observed across all models when utilizing D$^{s\rightarrow t}$ instead of D$^s$. This increase in metrics validates the proposed methodology of introducing FDA to curb the effect of adverse weather, pointing to a solution applicable to real-world scenarios.

The effect of the proposed methodology is also evident through the qualitative results, as observed in Fig. \ref{fig:FDA_comparisonl}. Looking closely, it is noted that models tend to detect traffic lights more accurately and with a much higher confidence score. Furthermore, on observing Fig. \ref{fig:light_instances}, it is noted through the Negative Samples (NS1, NS2) and Positive Samples (PS1, PS2) that the effect of FDA-augmented images extends to real-world scenarios where training on clear weather ($D^s$) may be inadequate and lead to missing detections. As evident in the figure, training on FDA-augmented images results in such traffic lights being detected with accurate bounding boxes. This exemplifies the applicability of the proposed methodology in real-world scenarios. In cases with insufficient data on traffic scenes in challenging weather conditions, FDA can be applied to existing data in clear environments. This can help the models generalize and perform better in unforeseen weather conditions. 

Referring to the SSL results in Table \ref{ssl_Table}, it is observed that training with only 50\% of the actual ground truth labels, while generating pseudo labels for the remaining data using YOLOv6, yields comparable performance. Evaluation was conducted using the same models as ones for fully supervised learning, i.e YOLOv5, YOLOv6, YOLOv8, and YOLOv10 on the NVIDIA A6000 for 20 epochs. Although the results may appear marginally suboptimal compared to fully supervised baselines, the implications are significant: this semi-supervised learning (SSL) approach demonstrates strong potential in scenarios where annotated data is inherently limited. Given the challenges in acquiring high-quality, annotated data in this domain—owing to both logistical and safety-related constraints—methods that effectively exploit limited supervision become essential. The results indicate that, even in the absence of 50\% of the original ground truth annotations, the proposed SSL-based pipeline achieves performance levels comparable to fully supervised models. This underscores the effectiveness of pseudo-labelling strategies in practical settings where annotated data is scarce and expensive to obtain. By effectively utilizing limited supervision, the method demonstrates significant potential for deployment in data-constrained environments.

However, it is essential to recognize certain limitations inherent to this approach. A primary concern is the potential reinforcement of dataset-specific biases introduced during pseudo-labelling, particularly when the model disproportionately learns from dominant patterns present in the limited labelled subset. Such biases may adversely affect generalization to unseen scenarios. Addressing this challenge remains an active area of investigation in our work, with ongoing efforts aimed at developing bias-aware training mechanisms to improve robustness and fairnes

s in model predictions.

\begin{table}[t]
    \small
    \setlength{\tabcolsep}{0.15\tabcolsep}
    \caption{Class-wise results when trained on D$^{s\rightarrow t}$ and inferred on the \(D^t\) (foggy and rainy) dataset. As evident across all metrics, the quantitative results of different models are impressive, especially for the Yellow images, showcasing the effectiveness of countering class imbalance.}
    \centering
    \label{table:classwise_results}
    \begin{tabular}{|c|c|c|c|c|c|c|}
    
        \hline
        \textbf{Model} & \textbf{$\beta$} & \textbf{Class} & \textbf{Precision} & \textbf{Recall} & \textbf{mAP50} & \textbf{mAP50-95} \\ \hline \hline
        YOLOv10n & 0.15 & Red & 0.938 & 0.893 & 0.952 & 0.698 \\
        ~ & ~ & Green & 0.912 & 0.809 & 0.91 & 0.623 \\
        ~ & ~ & Yellow & 0.882 & 0.847 & 0.918 & 0.529 \\
        \cline{2-7}
        ~ & 0.1 & Red & 0.942 & 0.871 & 0.94 & 0.669 \\
        ~ & ~ & Green & 0.895 & 0.802 & 0.887 & 0.586 \\
        ~ & ~ & Yellow & 0.874 & 0.817 & 0.883 & 0.485 \\
        \cline{2-7}
        ~ & 0.05 & Red & 0.949 & 0.941 & 0.973 & 0.749 \\
        ~ & ~ & Green & 0.922 & 0.898 & 0.952 & 0.687 \\
        ~ & ~ & Yellow & 0.934 & 0.931 & 0.966 & 0.609 \\
        \hline
        YOLOv8m & 0.15 & Red & 0.945 & 0.942 & 0.97 & 0.741 \\  
        ~ & ~ & Green & 0.922 & 0.888 & 0.942 & 0.678 \\ 
        ~ & ~ & Yellow & 0.934 & 0.939 & 0.957 & 0.603 \\ 
        \cline{2-7}
        ~ & 0.1 & Red & 0.949 & 0.899 & 0.952 & 0.704 \\
        ~ & ~ & Green & 0.915 & 0.835 & 0.9 & 0.629 \\ 
        ~ & ~ & Yellow & 0.919 & 0.898 & 0.929 & 0.565 \\ 
        \cline{2-7}
        ~ & 0.05 & Red & 0.908 & 0.71 & 0.811 & 0.563 \\
        ~ & ~ & Green & 0.856 & 0.664 & 0.738 & 0.491 \\ 
        ~ & ~ & Yellow & 0.883 & 0.703 & 0.751 & 0.431 \\ \hline
        YOLOv6m & 0.15 & Red & 0.934 & 0.89 & 0.934 & 0.662 \\ 
        ~ & ~ & Green & 0.899 & 0.813 & 0.895 & 0.588 \\ 
        ~ & ~ & Yellow & 0.908 & 0.865 & 0.914 & 0.506 \\ 
        \cline{2-7}
        ~ & 0.1 & Red & 0.932 & 0.817 & 0.89 & 0.599 \\
        ~ & ~ & Green & 0.878 & 0.732 & 0.82 & 0.519 \\ 
        ~ & ~ & Yellow & 0.859 & 0.769 & 0.818 & 0.432 \\ 
        \cline{2-7}
        ~ & 0.05 & Red & 0.952 & 0.936 & 0.968 & 0.730 \\ 
        ~ & ~ & Green & 0.913 & 0.895 & 0.94 & 0.658 \\ 
        ~ & ~ & Yellow & 0.931 & 0.926 & 0.953 & 0.574
        \\ \hline
        YOLOv5m & 0.15 & Red & 0.940 & 0.898 & 0.950 & 0.697 \\ 
        ~ & ~ & Green & 0.909 & 0.839 & 0.902 & 0.615 \\ 
        ~ & ~ & Yellow & 0.930 & 0.897 & 0.933 & 0.551 \\ 
        \cline{2-7}
        ~ & 0.1 & Red & 0.912 & 0.785 & 0.867 & 0.614 \\
        ~ & ~ & Green & 0.849 & 0.725 & 0.779 & 0.516 \\ 
        ~ & ~ & Yellow & 0.901 & 0.704 & 0.832 & 0.466 \\ 
        \cline{2-7}
        ~ & 0.05 & Red & 0.889 & 0.669 & 0.78 & 0.52 \\ 
        ~ & ~ & Green & 0.804 & 0.62 & 0.689 & 0.437 \\ 
        ~ & ~ & Yellow & 0.857 & 0.648 & 0.703 & 0.361
        \\ \hline \hline
    \end{tabular}
\end{table}

\begin{table}[]
    \centering
    \small
    \setlength{\tabcolsep}{0.5\tabcolsep}
    \caption{Comparisons of results of various models trained using SSL. As mentioned in the Benchmarking and Results section, pseudo-labels are generated for 50\% of D$^{s\rightarrow t}$ using YOLOv6m, while the rest are original labels.}
    \begin{tabular}{|c|c|c|c|c|c|}
    \hline
        \textbf{Model} & \textbf{Class} & \textbf{Precision} & \textbf{Recall} & \textbf{mAP50} & \textbf{mAP50-95} \\ \hline \hline
        YOLOv10n & All & 0.82 & 0.772 & 0.838 & 0.47 \\  \cline{2-6}
        ~ & Red & 0.816 & 0.832 & 0.876 & 0.54 \\ 
        \cline{2-6}
        ~ & Green & 0.786 & 0.718 & 0.805 & 0.455 \\ \cline{2-6}
        ~ & Yellow & 0.86 & 0.766 & 0.835 & 0.414    \\ 
        \hline
        YOLOv8m & All & 0.849 & 0.826 & 0.878 & 0.501 \\  \cline{2-6}
        ~ & Red & 0.842 & 0.876 & 0.908 & 0.565 \\ 
        \cline{2-6}
        ~ & Green & 0.808 & 0.771 & 0.833 & 0.478 \\ \cline{2-6}
        ~ & Yellow & 0.898 & 0.832 & 0.892 & 0.46 \\ 
        \hline
        YOLOv6m & All & 0.82 & 0.807 & 0.843 & 0.472 \\  \cline{2-6}
        ~ & Red & 0.861 & 0.862 & 0.89 & 0.543 \\ 
        \cline{2-6}
        ~ & Green & 0.79 & 0.751 & 0.81 & 0.455 \\ \cline{2-6}
        ~ & Yellow & 0.808 & 0.808 & 0.828 & 0.417 \\ 
        \hline
        YOLOv5m & All & 0.838 & 0.796 & 0.85 & 0.47 \\  \cline{2-6}
        ~ & Red & 0.828 & 0.854 & 0.887 & 0.532 \\ 
        \cline{2-6}
        ~ & Green & 0.795 & 0.745 & 0.8 & 0.441 \\ \cline{2-6}
        ~ & Yellow & 0.892 & 0.791 & 0.862 & 0.438 \\ 
        \hline \hline
    \end{tabular}
    \label{ssl_Table}
\end{table}

\section{Conclusion}
In this paper, we present a novel approach for Traffic Light Detection under Adverse Weather conditions, addressing two key challenges: domain shift and limited labelled data. To bridge the domain shift between clear and low-visibility scenarios such as rain and fog, we employ Fourier Domain Adaptation (FDA), which demonstrates improved metrics. Notably, as the size of the low frequency area ($\beta$) proportional to the image resolution being used for FDA increases from 0.05 to 0.15, model performance improves substantially, with $\beta = 0.15$ yielding the best and most consistent metrics. At this setting, YOLOv8 achieves the highest performance metrics: a Precision of 0.93, Recall of 0.92, mAP50 of 0.96, and mAP50-95 of 0.67—representing an average increase of $\mathbf{12.25\%}$ over the baseline. Furthermore, an average performance gain of $\mathbf{16.81\%}$ across all models is observed, including percentage improvements of 7.69\% in Precision, 19.91\% in Recall, 15.85\% in mAP50, and 23.81\% in mAP50-95. Semi-supervised learning (SSL) is implemented which demonstrates competitive performance with only 50\% of the available ground truth annotations used to generate pseudo-labels using YOLOv6m. To support rigorous evaluation, we curate a dedicated dataset comprising traffic light images across rainy and foggy conditions. These findings underscore the effectiveness of our approach in mitigating domain shift and label scarcity, highlighting its real-world applicability in challenging environmental conditions.


\section{Acknowledgment}

We would like to thank Mars Rover Manipal, an interdisciplinary student project of MAHE, for
providing the essential resources and infrastructure that supported our research.
\clearpage
\setcounter{page}{1}
\maketitlesupplementary

%

\begin{figure*}[t]
    \centering
    \small
    \includegraphics[width=\textwidth]{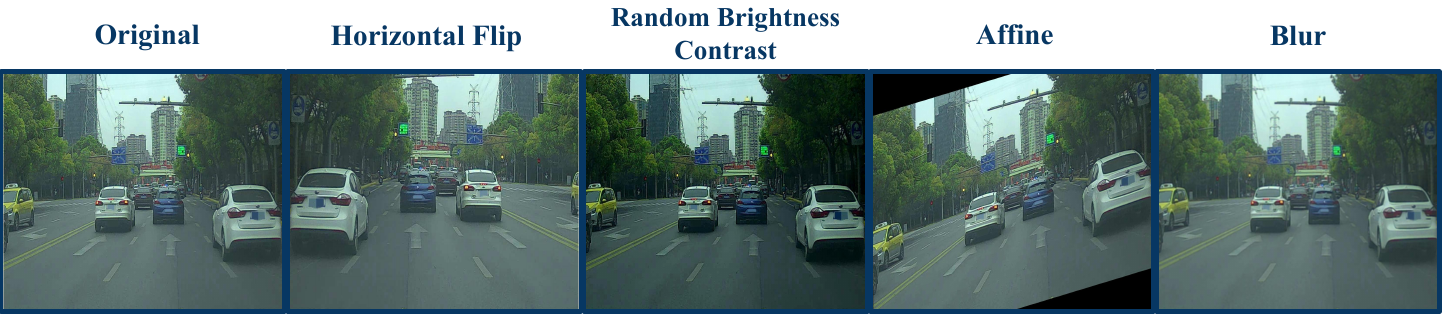}
    \caption{A visualization of the various methods used to augment images. The first image from the left is unaltered, the second is horizontally flipped, the third's contrast and brightness have been altered, the fourth has an affine transformation, and the last has been blurred. The hyperparameters for these augmentations are mentioned in Table \ref{table:aug_hyperparams} .}
    \label{fig:augmentations}
\end{figure*}

In this appendix, we provide supplementary material detailing the hyperparameters used in our experiments, along with an in-depth analysis of the extensive experimentation conducted to determine their optimal selection. Additionally, we present specifics of the benchmarking on clean images and tested on images affected by adverse weather conditions without applying FDA.

\subsection{Datasets}
The datasets LISA and S\textsuperscript{2}TLD were chosen since LISA supplies a rich night-time split, and S\textsuperscript{2}TLD  adds high-resolution images of very small lights under a permissive MIT licence. Together, they cover our key corner-cases, low-light scenes, and tiny targets, without extra relabelling or preprocessing. Whereas DriveU and Bosch BSTLD were discarded since, DriveU encodes 344 composite classes that combine colour with arrows, pictograms, and relevance flags; only a handful map directly to the simple colour states we need, so most would have to be merged or discarded. Bosch BSTLD, meanwhile, is converted from 12-bit sensor data to 8-bit RGB, a step that can introduce colour artefacts and risk degrading FDA, which depends on accurate pixel intensities.

\subsection{Hyperparameter Tuning}

\begin{table}[h]
    \centering
    \setlength{\tabcolsep}{0.5\tabcolsep}
    \small
    \caption{Comparisons of different models when trained on \(D^s\) (clean, $\beta$=0) data but inferred on the \(D^t\) (foggy and rainy) data.}
    \begin{tabular}{|c|c|c|c|c|c|}
    \hline
        \textbf{Model} & \textbf{Class} & \textbf{Precision} & \textbf{Recall} & \textbf{mAP50} & \textbf{mAP50-95} \\ \hline 
        \hline
        YOLOv10n & All & 0.86 & 0.691 & 0.774 & 0.46 \\ \cline{2-6}
        ~ & Red & 0.891 & 0.724 & 0.821 & 0.53 \\ \cline{2-6}
        ~ & Green & 0.873 & 0.651 & 0.745 & 0.445 \\ \cline{2-6}
        ~ & Yellow & 0.815 & 0.698 & 0.755 & 0.405 \\
        \hline
        YOLOv8m & All & 0.887 & 0.804 & 0.873 & 0.564 \\ \cline{2-6}
        ~ & Red & 0.937 & 0.825 & 0.91 & 0.641 \\ \cline{2-6}
        ~ & Green & 0.886 & 0.76 & 0.851 & 0.553 \\ \cline{2-6}
        ~ & Yellow & 0.839 & 0.828 & 0.859 & 0.499 \\ \hline
        YOLOv6m & All & 0.82 & 0.757 & 0.809 & 0.52 \\ \cline{2-6}
        ~ & Red & 0.914 & 0.793 & 0.877 & 0.61 \\ \cline{2-6}
        ~ & Green & 0.875 & 0.719 & 0.819 & 0.529 \\ \cline{2-6}
        ~ & Yellow & 0.671 & 0.758 & 0.729 & 0.422 \\
        \hline
        YOLOv5m & All & 0.856 & 0.682 & 0.766 & 0.481 \\ \cline{2-6}
        ~ & Red & 0.915 & 0.687 & 0.8147 & 0.549 \\ \cline{2-6}
        ~ & Green & 0.87 & 0.658 & 0.76 & 0.484 \\ \cline{2-6}
        ~ & Yellow & 0.784 & 0.7 & 0.722 & 0.41 \\
        \hline \hline
    \end{tabular}
    \label{table: trained_on_clean}
\end{table}

In order to increase the proportion of yellow images in the merged dataset, several augmentations were applied; as shown in Fig. \ref{fig:augmentations} and Table \ref{table:aug_hyperparams}, for the random brightness contrast adjustment, brightness and contrast limits were set to 0.2 (brightness\_limit=0.2, contrast\_limit=0.2) to modulate the image intensity dynamically. The affine transformation encompasses scaling, translation, rotation, and shearing. The affine transformation included a shear range set between 0 and 20 degrees (shear=(0, 20)). The blur augmentation had a blur limit with a kernel size 7 (blur\_limit=7).

\begin{figure}[!h]
    \centering
    \small
    \includegraphics[width=0.46\textwidth]{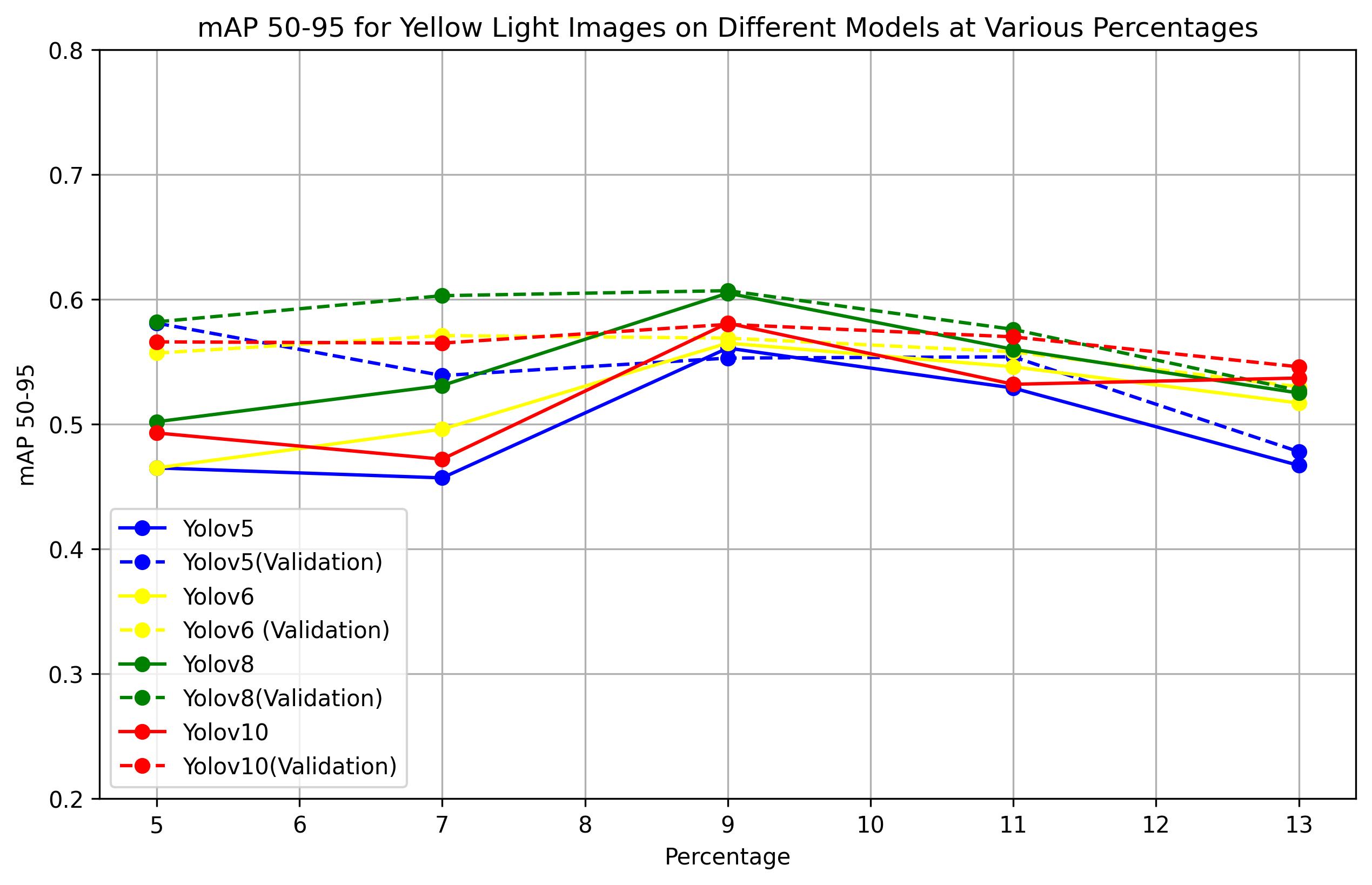}
    \caption{Benchmarked results of various models on various percentages of yellow lights. It is observed that all models converge at 13\%, indicating that the class imbalance is abated. Here, the percentage on the x-axis refers to the percentage of yellow lights in the merged dataset.}
    \label{fig:dataset_percentages}
\end{figure}

As discussed in Section \ref{sec:Experiments}, Fig. \ref{fig:dataset_percentages} illustrates findings which justify the choice of yellow lights as 13\%. At 9\%, YOLOv5 exhibits underfitting, as indicated by a noticeable gap between training and validation mAPs and overall lower performance compared to other models. This suggests that the limited data volume was insufficient for the model to capture the necessary distributional features, especially given its relatively smaller capacity compared to later versions like YOLOv8 and YOLOv10. Increasing to 13\% allowed YOLOv5 to generalize better and close the performance gap, justifying its use in our final training pipeline. The slight performance dip at 13\% is not necessarily indicative of model degradation. This behavior can arise due to increased data variance, introducing harder or noisier samples into the training set. The trade-off between marginal mAP fluctuations and improved real-world resilience is a well-documented phenomenon in deep learning. Therefore, selecting 13\% strikes a practical balance — it compensates for underfitting in lighter models like YOLOv5 and introduces diversity that benefits overall generalization, even if it momentarily lowers validation mAP on a fixed subset. Such effects are acceptable in deployment-focused applications where robustness outweigh

\begin{figure*}[t]
    \centering
    \includegraphics[width=\textwidth, height=3in, keepaspectratio]{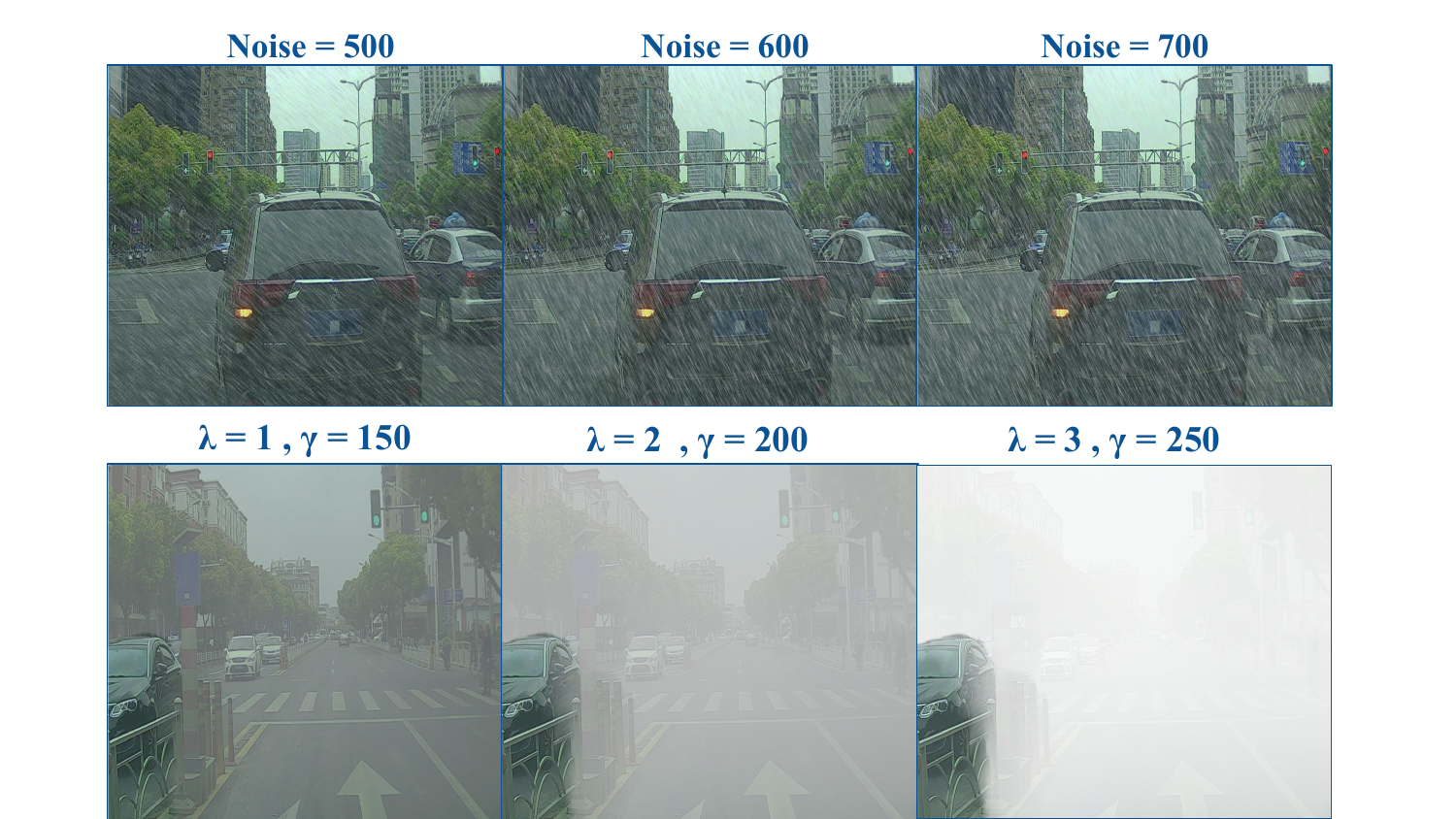}
i    \caption{The first row of images has rain added to them, and the second row has fog added to them. As evident from the images above, the set of hyperparameters, namely Noise=500, $\lambda$=1, and $\gamma$=150, result in a realistic and balanced effect.}
    \label{fig:Hyperparams_examples}
\end{figure*}

As mentioned in Section \ref{sec:Experiments}, the hyperparameters employed for addition of fog are $\lambda$ and Airlight ($\gamma$). The amount of fog in an image is determined by $\lambda$, while $\gamma$ controls the brightness of the foggy image. In the case of rain addition, the hyperparameters include noise, rain\_len, rain\_angle, rain\_thickness, and $\alpha$. Noise determines the density of rain in the image; rain\_len specifies the length of each raindrop; rain\_angle dictates the angle of the raindrops relative to the vertical axis; rain\_thickness defines the thickness of each raindrop; and $\alpha$ sets the opacity of the raindrops.
After extensive experimentation, we believe the suitable set of hyperparameters to be $\lambda = \text{1}$, $\gamma = \text{150}$, noise = \text{500}, rain\_len = \text{[50,60]}, rain\_angle = \text{[-50,51]}, rain\_thickness = \text{3}, and $\alpha$ = \text{0.7}.
Examples of these hyperparameters with different values can be seen in Fig. \ref{fig:Hyperparams_examples}.

\subsection{Benchmarking}
This section provides a summary of the different YOLO models used in the experiments, highlighting their architectural differences and unique characteristics. 


\begin{table}[t]
    \centering
    \small
    \label{table:augmentations_table}
    \caption{The list of augmentations applied to address the class imbalance and the hyperparameters employed in the process. All these hyperparameters were used with a probability of 0.5.}
    \begin{tabular}{|c|c|}
    \hline
       \textbf{Augmentation} & \textbf{Hyperparameters} \\ \hline
       \hline
        Horizontal Flip & p=0.5 \\ \hline
        Random Brightness Contrast & brightness\_limit=0.2, \\ 
        ~ & contrast\_limit=0.2 \\ \hline
        Affine & shear=(0,20) \\ \hline
        Blur & blur\_limit=7 \\ \hline \hline
    \end{tabular}
    \label{table:aug_hyperparams}
\end{table}

\textbf{YOLOv5} starts with a strided convolution layer to reduce memory and computational costs, and the SPPF (Spatial Pyramid Pooling Fast) layer accelerates computation by pooling multi-scale features into a fixed-size feature map.

\textbf{YOLOv6} introduced a new backbone called EfficientRep \cite{weng2023efficientrep}, built on RepVGG \cite{ding2021repvggmakingvggstyleconvnets}, which offered greater parallelism than previous YOLO backbones. The network's neck used a PAN (Path Aggregation Network) structure, enhanced with RepBlocks or CSPStackRep Blocks for larger models. This redesigned backbone and neck significantly improved the model's efficiency and adaptability.

\textbf{YOLOv8} retains a backbone architecture similar to YOLOv5 but introduces significant changes to the CSPLayer, now called the C2f module. This module, which stands for "cross-stage partial bottleneck with two convolutions," effectively merges high-level features with contextual information, resulting in improved detection accuracy. 

\textbf{YOLOv10} introduces a lightweight classification head with depth-wise separable convolutions to reduce computational overhead, Spatial-Channel Decoupled Downsampling to minimize information loss, and Rank-Guided Block Design for optimal parameter use. Accuracy is enhanced through Large-Kernel Convolution for better feature extraction and Partial Self-Attention (PSA) for improved global representation with minimal overhead. It also uses dual label assignments with a consistent matching metric, allowing for rich supervision and efficient deployment without the need for NMS, an integral component of previous YOLO versions. 

Each of these YOLO versions differs in its architectural innovations and optimizations, catering to different aspects of object detection challenges. All aforementioned models were trained using the Adam optimizer 
\cite{kingma2017adammethodstochasticoptimization} for 50 epochs on the NVIDIA RTX A6000 GPU with a batch size of 32.

The models are benchmarked by training them on the clean images without rain or fog and testing them on the images that have rain/fog in them.  The results of this benchmarking process, which highlight the impact of training on clean images and testing on weather-degraded images, are presented in Table \ref{table: trained_on_clean}.

\subsection{Additional Results}
In Tables \ref{table: trained_on_clean} and \ref{table: aggregate_fda_results}, two scenarios are presented: (1) when models are trained on $D\textsuperscript{s}$ but tested on $D\textsuperscript{t}$, and (2) when models are trained on $D^{s \to t}$ and evaluated on $D\textsuperscript{t}$, for which the subplots are shown for representational purposes in the main paper in Fig. \ref{fig:results_graph} and the exact values for which are mentioned in Table \ref{table: aggregate_fda_results}.

\begin{table}[t]
    \small
    \setlength{\tabcolsep}{0.4\tabcolsep}
    \centering
    \caption{Comparisons of results across models when trained on D$^{s\rightarrow t}$ and inferred on $D^t$ (rainy and foggy). All metrics are averaged across the 3 classes.}
    \label{table: aggregate_fda_results}
    \begin{tabular}{|c|c|c|c|c|c|}
    \hline
        \textbf{Model} & \textbf{Beta($\beta$)} & \textbf{Precision} & \textbf{Recall} & \textbf{mAP50} & \textbf{mAP50-95} \\ \hline \
        YOLOv10n & 0.15 & 0.911 & 0.85 & 0.927 & 0.617 \\
        ~ & 0.1 & 0.903 & 0.83 & 0.903 & 0.58 \\
        ~ & 0.05 & 0.935 & 0.923 & 0.964 & 0.682 \\
        \hline
        YOLOv8m & 0.15 & 0.933 & 0.923 & 0.956 & 0.674 \\  
        ~ & 0.1 & 0.927 & 0.877 & 0.927 & 0.633 \\ 
        ~ & 0.05 & 0.882 & 0.693 & 0.767 & 0.495 \\ \hline
        YOLOv6m & 0.15 & 0.914 & 0.856 & 0.914 & 0.585 \\ 
        ~ & 0.1 & 0.890 & 0.772 & 0.843 & 0.516 \\
        ~ & 0.05 & 0.932 & 0.919 & 0.954 & 0.654 \\
        \hline
        YOLOv5m & 0.15 & 0.926 & 0.878 & 0.928 & 0.621 \\ 
        ~ & 0.1 & 0.887 & 0.765 & 0.826 & 0.532 \\
        ~ & 0.05 & 0.850 & 0.646 & 0.724 & 0.439 \\\hline \hline
    \end{tabular}
\end{table}

{
    \small
    \bibliographystyle{ieeenat_fullname}
    \bibliography{main}
}

\end{document}